\documentclass[letterpaper, 10 pt, conference]{ieeeconf}
\IEEEoverridecommandlockouts

\usepackage{tcolorbox}
\usepackage{xcolor}
\usepackage{cite}
\usepackage{amsmath,amssymb,amsfonts}
\usepackage{algorithmic}
\usepackage{graphicx}
\usepackage{textcomp}
\usepackage{xcolor}
\usepackage{lipsum}
\usepackage{booktabs}
\usepackage{placeins}
\usepackage{afterpage}
\usepackage{colortbl}
\usepackage[hidelinks]{hyperref}

\FloatBarrier
\definecolor{mygreen}{RGB}{0, 0, 0}
\def\BibTeX{{\rm B\kern-.05em{\sc i\kern-.025em b}\kern-.08em
    T\kern-.1667em\lower.7ex\hbox{E}\kern-.125emX}}
    
\begin{document}

\title{\LARGE \bf
NutriGen: Personalized Meal Plan Generator Leveraging Large Language Models to Enhance Dietary and Nutritional Adherence
}

\author{Saman Khamesian$^{1,2}$, Asiful Arefeen$^{1,2}$, Stephanie M. Carpenter$^{1}$, and Hassan Ghasemzadeh$^{1}$
\thanks{$^{1}$College of Health Solutions, Arizona State University, Phoenix, AZ 85004, USA.}
\thanks{$^{2}$School of Computing and Augmented Intelligence, Arizona State University, Tempe, AZ 85281, USA.}
\thanks{{Corresponding author: \textcolor{blue}{skhamesi@asu.edu}}}}

\maketitle

\begin{abstract}
Maintaining a balanced diet is essential for overall health, yet many individuals struggle with meal planning due to nutritional complexity, time constraints, and lack of dietary knowledge. Personalized food recommendations can help address these challenges by tailoring meal plans to individual preferences, habits, and dietary restrictions. However, existing dietary recommendation systems often lack adaptability, fail to consider real-world constraints such as food ingredient availability, and require extensive user input, making them impractical for sustainable and scalable daily use. To address these limitations, we introduce \textbf{NutriGen}, a framework based on large language models (LLM) designed to generate personalized meal plans that align with user-defined dietary preferences and constraints. By building a personalized nutrition database and leveraging prompt engineering, our approach enables LLMs to incorporate reliable nutritional references like the USDA nutrition database while maintaining flexibility and ease-of-use. We demonstrate that LLMs have strong potential in generating accurate and user-friendly food recommendations, addressing key limitations in existing dietary recommendation systems by providing structured, practical, and scalable meal plans. Our evaluation shows that Llama 3.1 8B and GPT-3.5 Turbo achieve the lowest percentage errors of 1.55\% and 3.68\%, respectively, producing meal plans that closely align with user-defined caloric targets while minimizing deviation and improving precision. Additionally, we compared the performance of DeepSeek V3 against several established models to evaluate its potential in personalized nutrition planning. Our results showed that it struggled with accuracy, exhibiting a higher MAE of 10.45\%, and was significantly slower in processing time compared to models like GPT-3.5 Turbo and Llama 3.1 8B, highlighting the need for further optimization.
\newline

\end{abstract}

\section{INTRODUCTION}
Maintaining a healthy diet is fundamental to preventing and managing a range of chronic diseases, including obesity, cardiovascular conditions, and diabetes \cite{brownson2004chronic}. Despite its importance, many individuals face significant challenges in adhering to a healthy eating regimen. These difficulties often stem from the need to calculate daily caloric intake, evaluate the nutritional value of meals, and consistently make informed decisions about what to eat. Additionally, planning balanced meal schedules or preparing meals with the ingredients available at home can be overwhelming, especially when trying to align with specific dietary goals and restrictions. These barriers highlight the pressing need for tools that simplify these decision making processes, making it easier for individuals to adopt and sustain healthier eating habits over time \cite{alshurafa2015recognition}.

To address the aforementioned challenges, personalized dietary recommendation systems have emerged as a promising solution. Unlike conventional food recommendation systems, which primarily consider whether users enjoy specific foods, personalized approaches cater to both individual dietary preferences and health needs, such as gluten-free diets, allergy sensitivities, and vegetarian lifestyles \cite{farquhar1977community}. These systems aim to meet unique nutritional requirements, ensuring better adherence and promoting long-term health outcomes \cite{kim2020knowledge, arefeen2022computational}. Furthermore, nutrition-oriented food recommendations go beyond basic food preferences by integrating the broader impacts of dietary patterns on individual health. By aligning personal preferences with evidence-based nutritional guidelines, these systems offer actionable insights to help users make informed and health-conscious dietary decisions \cite{yang2024chatdiet}.

The rise of machine learning (ML) and data analysis techniques has further enhanced the effectiveness of personalized dietary recommendations \cite{sherwood2016bestfit, hezarjaribi2019human, hezarjaribi2017speech2health}. By leveraging user data, these advanced systems can generate insights into optimal eating patterns while considering individual preferences. For example, DietOS employs optimization algorithms to develop customized diet plans tailored to the management of chronic diseases \cite{agapito2018dietos}. Similarly, IoMT-assisted systems, such as the approach by Iwendi et al., utilize ML to create precise, patient-specific dietary recommendations, excelling in clinical environments \cite{iwendi2020realizing}. ChatDiet integrates large language models (LLM) to deliver interactive and engaging dietary advice via chatbot interfaces, significantly enhancing user interaction \cite{yang2024chatdiet}. Additionally, optimization techniques like Particle Swarm Optimization (PSO) have been applied to generate meal plans that efficiently balance nutritional goals \cite{narendra2023particle}. These innovative methodologies illustrate the potential of computational tools in addressing dietary challenges.

\begin{figure*}[!tb]
    \centerline{\includegraphics[scale=0.35]{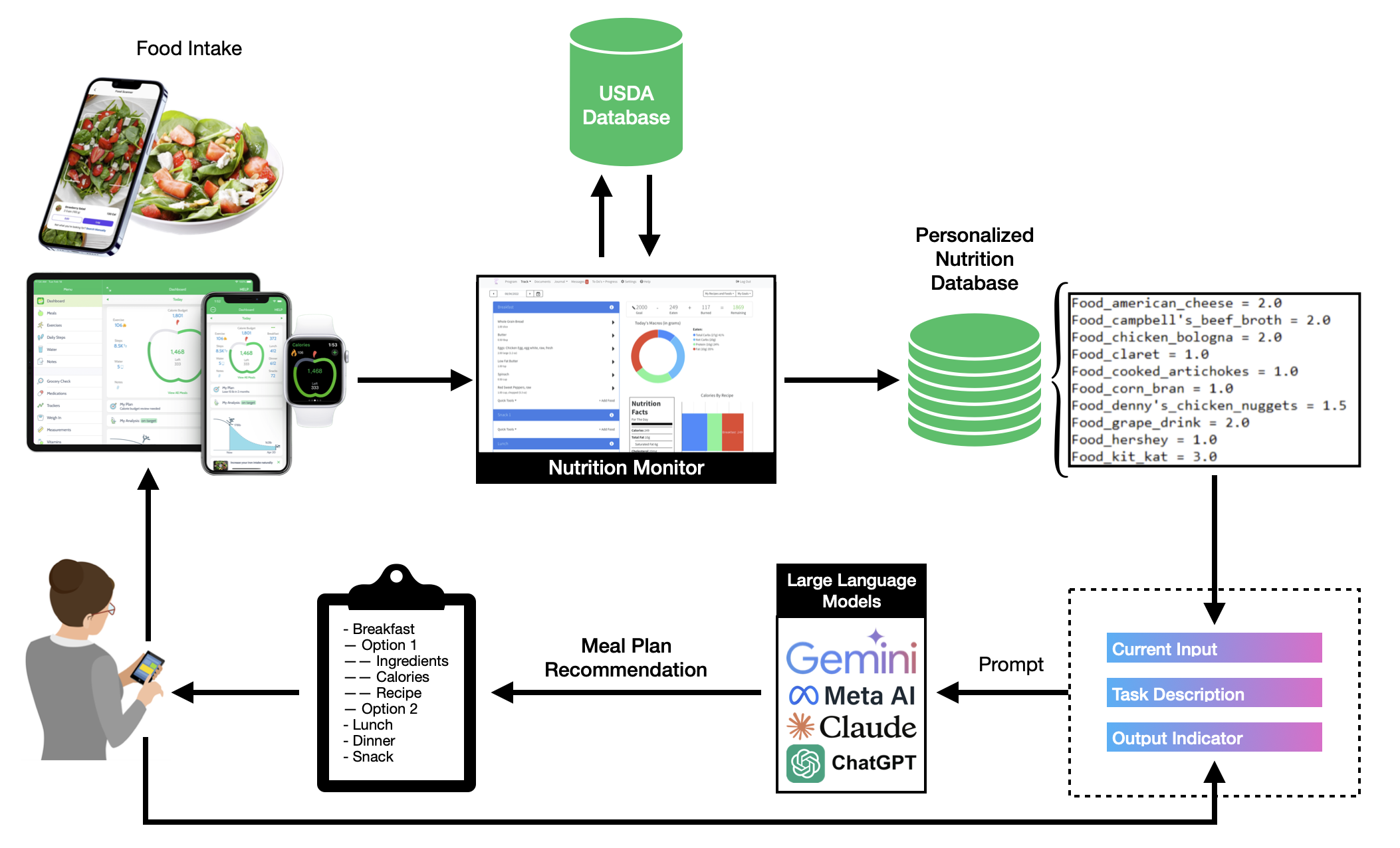}}
    \caption{The overall architecture of the NutriGen framework for personalized meal plan recommendations. The system integrates user inputs, external nutrition databases, and LLMs to generate structured and practical meal plans.}
    \label{fig:methodology}
\end{figure*}

However, despite their potential, existing dietary recommendation systems face notable limitations that restrict their practical application. Methods such as PSO, while efficient at balancing nutritional objectives, often fail to account for user-specific preferences like cultural restrictions or diverse dietary needs, which reduces their real-world usability \cite{narendra2023particle}. Similarly, IoMT-assisted systems, though highly precise and robust in clinical settings, are not well-suited for the general population seeking practical, everyday dietary guidance. Their reliance on continuous medical monitoring, clinical validation, and specialized sensors limits their usability outside structured healthcare environments, reducing their practicality for everyday meal planning \cite{iwendi2020realizing}. ChatDiet excels in user engagement but struggles with scalability due to high computational costs, platform dependencies, and reliance on continuous user input \cite{yang2024chatdiet}. Moreover, many existing systems focus solely on recommending individual food items rather than offering comprehensive and cohesive diet menus. This leaves users with the burden of assembling balanced meal plans on their own. Additionally, these systems frequently lack features such as alternative food suggestions, consideration of ingredient availability, or integrated recipes and cooking instructions. As a result, users may find it challenging to implement the recommendations in their daily lives. These challenges often lead to steady declines in engagement over time, with many participants discontinuing their use of the tools altogether. Addressing these gaps requires the development of adaptable, user-centered frameworks that provide structured, actionable, and personalized dietary guidance \cite{wang2024optimization, khamesian2025type, story2008creating, mamun2024use}.

In this paper, we introduce NutriGen, a personalized meal plan recommendation system designed to overcome the limitations of existing methods by generating comprehensive and user-friendly daily meal plans. The system offers multiple meal plans, each including options for breakfast, lunch, dinner, and snacks, along with calorie counts, nutritional information and recipes. By leveraging LLMs and prompt engineering, the framework incorporates user preferences, restrictions, and ingredient availability to deliver fully personalized and practical dietary recommendations. This approach bridges the gap between nutrition science and usability, offering a flexible solution to promote sustainable healthy eating habits.

\section{METHODOLOGY}
In this section, we present our NutriGen framework for generating personalized meal plans based on users' food intake history, habits, preferences, and dietary restrictions. Existing dietary recommendation systems often lack adaptability, failing to accommodate diverse user needs and real-world constraints. NutriGen addresses these limitations by leveraging LLMs and prompt engineering to integrate user-specific inputs into a structured nutrition database. This approach ensures that meal plans align with individual dietary goals while incorporating features such as calorie counts, alternative food options, and suggested recipes. These enhancements improve the practical usability of personalized dietary recommendations \cite{wu2024survey, li2024pre, sahoo2024systematic}. The overall architecture of NutriGen is illustrated in Fig.~\ref{fig:methodology}.

\subsection{Input Data}
The system collects input from multiple sources, including food trackers that utilize image capture, manual text entry, and voice input. Image-based food logging employs optical character recognition (OCR) and ML models to estimate portion sizes and extract nutritional details, automating the process and reducing user effort. Text and voice-based inputs leverage natural language processing (NLP) to interpret descriptions, enabling flexible and user-friendly data entry. Together, these methods create a comprehensive record of a user's dietary intake, improving the accuracy and relevance of generated meal plans \cite{yang2017yum}.

To enhance reliability, user-reported data is supplemented with standardized nutritional values retrieved from external sources, such as the United States Department of Agriculture (USDA) database\footnote{\href{https://fdc.nal.usda.gov/}{https://fdc.nal.usda.gov/}}. Cross-referencing with established dietary guidelines helps mitigate potential inaccuracies in manual entries while ensuring consistency in nutrient estimations. Additionally, the system can integrate multiple data repositories, including branded food databases and nutrition labels, allowing for greater adaptability across different dietary contexts. Over time, user interactions, preferences, and feedback are incorporated into the nutrition database, refining recommendations dynamically to better align with individual dietary goals.

\subsection{Architecture}
Once the personalized nutrition database is constructed, it serves as the foundation for prompt engineering. The input prompt for the language model is structured into three components: the current input, the task description, and the output indicator. The structured prompt \( P \) is formally defined as:

\[P = \{ I_{\text{current}}, T_{\text{task}}, O_{\text{output}} \}\]

\noindent where:

\begin{itemize}
    \item \( I_{\text{current}} \) represents the user's dietary profile, including food intake history, preferences, and constraints.
    \item \( T_{\text{task}} \) defines the task with structured instructions, such as generating meal plans that meet specific nutritional targets (e.g., calorie goals, macronutrient distribution, dietary restrictions).
    \item \( O_{\text{output}} \) specifies the expected format, ensuring clarity and adherence to user-defined requirements.
\end{itemize}
Using this structured prompt, the meal plan generation process is formulated as:

\[M_i = \text{LLM}(P)\]
This approach enables self-contained meal generation without external dependencies, ensuring that outputs are consistent, personalized, and aligned with user-defined goals.

A potential extension to improve the accuracy of nutritional estimations in meal plan generation is the Retrieval-Augmented Generation (RAG) framework. RAG integrates a retriever module, \( Q(x) \), which queries an external knowledge base such as the USDA database for a given meal \( x \), with a generator module, \( G(Q(x)) \), powered by an LLM. This approach allows the system to dynamically retrieve up-to-date nutritional values, reducing discrepancies between estimated and actual calorie counts. Incorporating RAG into the system would allow the meal plan generation process to be formalized as:

\[M_i = G(Q(x), P) = \text{LLM}(Q(x), P)\]

\noindent where:

\begin{itemize}
    \item \( Q(x) \) retrieves the USDA-referenced nutritional values for meal \( x \), ensuring accuracy.
    \item \( G(Q(x)) \) generates a meal plan that aligns with the retrieved nutritional information and user specifications.
    \item \( P \) serves as the structured prompt, integrating user preferences and constraints into the generation process.
\end{itemize}

While RAG offers the advantage of retrieving real-time USDA nutritional values, it also introduces practical challenges such as increased latency, token limits, and dependency on external APIs. Future work may explore hybrid approaches, where retrieved nutritional data is pre-processed and integrated into the structured prompt before querying the LLM.

\subsection{Prompt Engineering}
In designing the prompt, we structured it to ensure meal plans are personalized, nutritionally balanced, and aligned with user-defined goals. The prompt accommodates diverse dietary preferences, restrictions (e.g., gluten-free, vegetarian), and specific nutritional targets by explicitly defining constraints such as calorie limits, macronutrient distributions, and portion sizes. To maintain consistency and clarity across model outputs, we employed a modular structure, where different sections progressively refine the generated meal plans.

To further enhance reliability, we incorporated few-shot prompting by providing structured examples within the prompt. These examples serve as reference points, guiding the LLM to follow a consistent format while allowing diversity in meal recommendations. Additionally, our prompt design supports flexible constraints, enabling users to prioritize specific aspects such as high-protein meals or low-sugar alternatives. Below is the core section of the structured template prompt used in all our tests. The complete prompt and implementation code \footnote{\href{https://github.com/SamanKhamesian/NutriGen}{https://github.com/SamanKhamesian/NutriGen}} is available for reference.

\begin{table}[htbp!]
    \centering
    \begin{tcolorbox}[colframe=black, colback=white, sharp corners=south, boxrule=0.8pt]
        \textcolor{mygreen}{\textbf{\ttfamily
        User's goal is to create a meal plan with: \\
        - Total calories: \{total\_calories\} kcal. \\
        - Total protein: \{target\_protein\} g. \\
        - Total sugar: \{target\_sugar\} g. \\
        The plan must include: \\
        - Breakfast, lunch, dinner, and snacks. \\
        - The calorie count for each meal (e.g., "Breakfast: 400 kcal"). \\
        - At the end of each meal plan option, provide the total calories, total fat, total protein, and total carbohydrate. \\
        - For each meal component, specify portion sizes, including the number of items or volume (e.g., "1 Kit Kat bar (45g)," "1 hamburger with a 150g beef patty, bun, and lettuce"). \\
        - Provide a short recipe for each item, detailing how it can be prepared (e.g., "Grill the patty for 5 minutes, then assemble with lettuce, tomato, and a bun"). \\
        Provide three different meal plan options for diversity. \\
        Use familiar dishes instead of listing individual food items. For example, use "hamburger" instead of "150 grams of meat with bun and lettuce." \\
        Ensure the plan adheres to the user's preferences and restrictions and meets the specified targets while maintaining a balanced nutritional profile. \\
        Here are the available items: \{menu\_input\}
        }}
    \end{tcolorbox}
    \label{tab:input_prompt}
\end{table}

\section{EXPERIMENTAL SETUP}
\subsection{Dataset}
The input prompt requires information on the user’s consumption patterns, target calorie intake, and intervention duration as inputs to the system. To generate meal plans using LLMs without conducting a user study, we simulated realistic initial consumption patterns. Instead of collecting dietary data from human participants, we designed a system to generate ten diverse and plausible daily food intake profiles using a set of 200 randomly selected meals from the USDA dataset. To introduce variability and ensure a broad representation of dietary scenarios, calorie goals and intervention durations were randomly selected from a reasonable range (1500–3500 kcal). These synthetic consumption patterns, along with corresponding diet goals and durations, were then incorporated into input prompts to generate personalized meal plan recommendations. This approach allowed us to bypass the logistical and ethical challenges associated with human data collection while still evaluating the system’s ability to tailor recommendations. The following box presents a sample nutrition profile with predefined food items and nutritional targets. These values were used as input for the LLMs, which were then tasked with generating meal plans that align with these specifications.

\begin{table}[htbp!]
    \centering
    \begin{tcolorbox}[colframe=black, colback=white, sharp corners=south, boxrule=0.8pt]
    \textcolor{mygreen}{\textbf{\ttfamily
        Food\_Barbequeue\_Lays = 1.0 \\
        Food\_Garden\_Pizza = 1.0 \\
        Food\_Milano\_double\_chocolate = 1.0 \\
        Food\_baked\_cheddar\_ruffles = 1.0 \\
        Food\_beef\_angus\_burger\_patty = 1.0 \\
        Food\_chocolate\_milkshake = 1.0 \\
        Food\_eggs\_benedict = 0.5 \\
        Food\_tortilla\_chips = 1.0 \\
        \noindent\rule{8cm}{0.4pt} \\
        calories = 1573.25 \\
        protein = 54.0 \\
        sugar = 58.3
        }}
    \end{tcolorbox}
    \label{tab:nutrition_profile}
\end{table}

\subsection{Selected Models}
In our experimental evaluations, we selected several advanced language models from leading organizations:

\begin{itemize}
    \item \textbf{Llama 3.1-70B \cite{grattafiori2024llama}}: Developed by Meta, this model comprises 70 billion parameters and has been trained on approximately 15 trillion tokens, demonstrating enhanced performance in language understanding and generation tasks.
    \item \textbf{Llama 3.1-8B \cite{grattafiori2024llama}}: A smaller variant from Meta, this 8 billion parameter model is designed for efficiency while maintaining robust language processing abilities.
    \item \textbf{Claude 3.5 Sonnet}: Anthropic's model, with an undisclosed number of parameters, emphasizes safety and reliability, exhibiting improved capabilities in coding and reasoning tasks.
    \item \textbf{Claude 3.5 Haiku}: A streamlined version focusing on rapid response generation and concise outputs, tailored for applications requiring brevity and speed.
    \item \textbf{DeepSeek-V3 \cite{liu2024deepseek}}: Released by DeepSeek in December 2024, this model features 671 billion active parameters and a context length of 128K tokens, excelling in multilingual tasks, particularly in English and Chinese.
    \item \textbf{Gemini 2.0 Flash Exp \cite{team2023gemini}}: Developed by Google DeepMind, this model combines language understanding with reinforcement learning to enhance reasoning and problem-solving.
    \item \textbf{Gemini 1.5 Pro \cite{team2023gemini}}: Offers advanced language understanding and reasoning capabilities, optimized for professional application.
    \item \textbf{GPT-4o \cite{hurst2024gpt}}: OpenAI's latest model supports both text and image inputs, maintaining high intelligence levels across various tasks.
    \item \textbf{GPT-4o Mini}: A smaller model designed for high performance with lower resource usage.
    \item \textbf{GPT-3.5 Turbo \cite{ye2023comprehensive}}: An optimized version of GPT-3, offering faster response times and improved efficiency in language tasks.
\end{itemize}

These models were chosen to provide a comprehensive assessment of current advancements in language model development. In selecting the models, we aimed to include at least two from each model family—one representing the latest and most powerful model and another optimized for speed and efficiency. This approach ensures that our evaluation covers a diverse range of models, from large-scale, high-performance architectures to lightweight, fast alternatives, thereby capturing the full spectrum of capabilities available in modern language models.

\begin{figure}[b]
    \centering
    \includegraphics[width=0.45\textwidth]{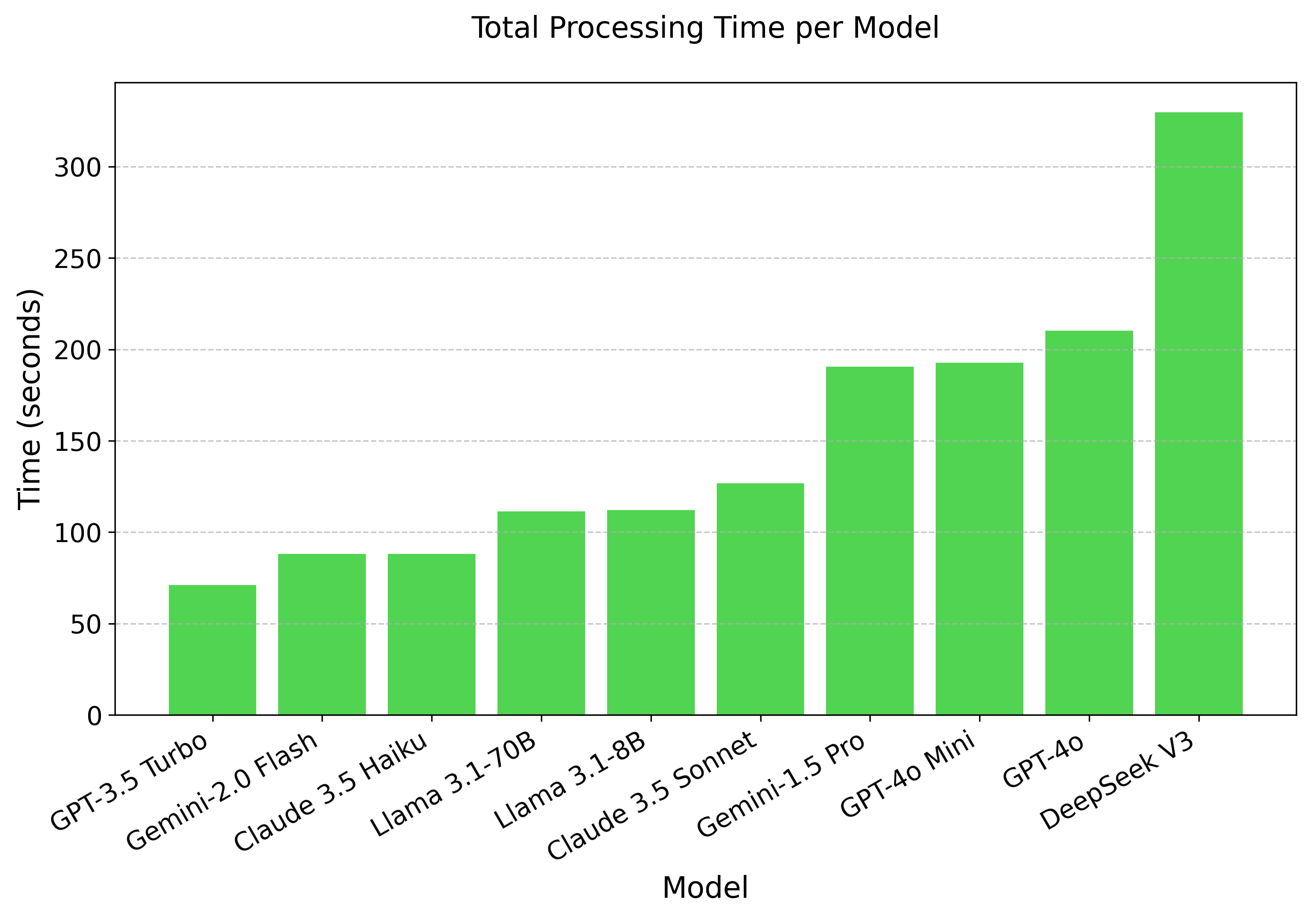}
    \caption{Comparison of total processing time}
    \label{fig:time_comparison}
\end{figure}

\section{RESULTS}
\subsection{Processing Time Comparison}

\begin{table*}[t]
    \centering
    \caption{Evaluation of Nutrition Facts Generated by LLMs: This table presents the error between the average reported total calories and the USDA database for various meal plans across different models. Bold numbers indicate the minimum error, while a dash (-) signifies that the model failed to generate a meal plan.}
    \label{tab:error_between_reported_nutrition_and_usda}
    \setlength{\tabcolsep}{0.9em}
    \renewcommand{\arraystretch}{1.1}
    \begin{tabular}{lcccccccccc}
        \toprule
        & \textbf{Input 1} & \textbf{Input 2} & \textbf{Input 3} & \textbf{Input 4} & \textbf{Input 5} & \textbf{Input 6} & \textbf{Input 7} & \textbf{Input 8} & \textbf{Input 9} & \textbf{Input 10} \\
        \midrule
        Claude 3.5 Haiku  & 570.69 & 124 & 88 & 103.5 & - & 75 & 76.75 & 100 & 83 & 113 \\
        Claude 3.5 Sonnet  & 138.09 & 124 & 88 & 105 & 96 & 75 & 77 & 96 & 83 & 113 \\
        DeepSeek V3 \cite{liu2024deepseek}      & 138.09 & 96.6 & 88 & - & - & - & - & 100 & 82.84 & 113 \\
        Gemini 1.5 Pro \cite{team2023gemini}    & 588 & 173.59 & 87.75 & 153.5 & 445.5 & 75 & 76.75 & 96 & 83 & 52.59 \\
        Gemini 2 Flash \cite{team2023gemini}    & 589.33 & 173.59 & 87.75 & 153.5 & 445.5 & 75 & 103.41 & 96 & 83 & 48 \\
        GPT-3.5 Turbo \cite{ye2023comprehensive}    & 138.09 & 73.59 & 37.75 & 53.5 & \textbf{45.5} & \textbf{25} & \textbf{26.75} & 46 & \textbf{32.84} & 62.59 \\
        GPT-4o \cite{hurst2024gpt}          & 50 & \textbf{24} & \textbf{38} & \textbf{45} & 563 & 41.6 & 50 & \textbf{36} & 74.6 & 52.59 \\
        GPT-4o Mini      & 60.03 & 27.5 & 50 & 50 & 161.66 & 55 & \textbf{26.75} & 47.33 & 35 & 53.33 \\
        Llama 3.1 8B \cite{grattafiori2024llama}     & 88.09 & 58.33 & 50 & 60 & 46 & 58.33 & \textbf{26.75} & - & 50 & \textbf{43.33} \\
        Llama 3.1 70B \cite{grattafiori2024llama}    & \textbf{45} & 60 & 50 & 50 & 46.66 & 40 & 48.33 & 46 & 50 & 63 \\
        \bottomrule
    \end{tabular}
\end{table*}

To evaluate the computational efficiency of each model, we measured the total processing time required to generate 10 outputs. The results, presented in Fig.~\ref{fig:time_comparison}, show significant variations in execution time across different models. As expected, smaller and optimized models such as \textbf{GPT-3.5 Turbo, Gemini 2.0 Flash, and Claude 3.5 Haiku} demonstrated the fastest processing times, completing the task in significantly less time compared to larger models.

In contrast, larger models required approximately twice the time to generate outputs. DeepSeek V3 exhibited the longest runtime, indicating higher computational demands. The comparison also highlights that while certain high-performance models, such as GPT-4o and Gemini 1.5 Pro, provide advanced capabilities, they require considerably more time to process outputs. These findings emphasize the trade-offs between model size, computational efficiency, and response time, which are critical considerations for real-time applications.

\subsection{Reported Nutritional Values vs. USDA Dataset}
One crucial aspect of our analysis is verifying the accuracy of nutrition facts generated by LLMs. Each model was asked to generate three meal plans for each input, and we computed the average absolute error between the total reported calories and the USDA reference values. When we ask a model to create a meal plan and compute total nutritional values, such as calories, fats, and protein, it is essential to assess whether the provided information is accurate. To validate these outputs, we used the USDA database as a reference and examined whether the nutritional values generated for meal plan items align with the actual USDA values. The Mean Absolute Error (MAE) was calculated as follows:

\begin{equation}
MAE = \frac{1}{M} \sum_{j=1}^{M} \left| \sum_{k=1}^{P} c_{\text{reported}, j, k} - \sum_{k=1}^{P} c_{\text{USDA}, j, k} \right|
\end{equation}
where \( M = 3 \) represents the number of meal plans generated per input, and \( P \) is the number of meal items in each meal plan. Here, \( c_{\text{reported}, j, k} \) denotes the reported calorie value of item \( k \) in meal plan \( j \), while \( c_{\text{USDA}, j, k} \) represents the corresponding USDA reference value. This formula ensures that the total calories for each meal plan are first computed by summing up the calorie values of individual meal items. Then, the absolute difference between the reported total and the USDA reference is calculated for each meal plan. The average across the three meal plans provides a robust measure of the model’s accuracy in estimating calorie values.

Table \ref{tab:error_between_reported_nutrition_and_usda} presents the error between the average reported total calories per input and the corresponding USDA values. The results demonstrate that \textbf{GPT-4o} and \textbf{GPT-3.5 Turbo} achieve the lowest error, alongside other models that exhibit their capability to retrieve accurate nutritional information from reliable sources.

\subsection{Adherence of Meal Plans to User-Specified Targets}
\begin{figure*}[t]
    \centering
    \includegraphics[width=1\textwidth]{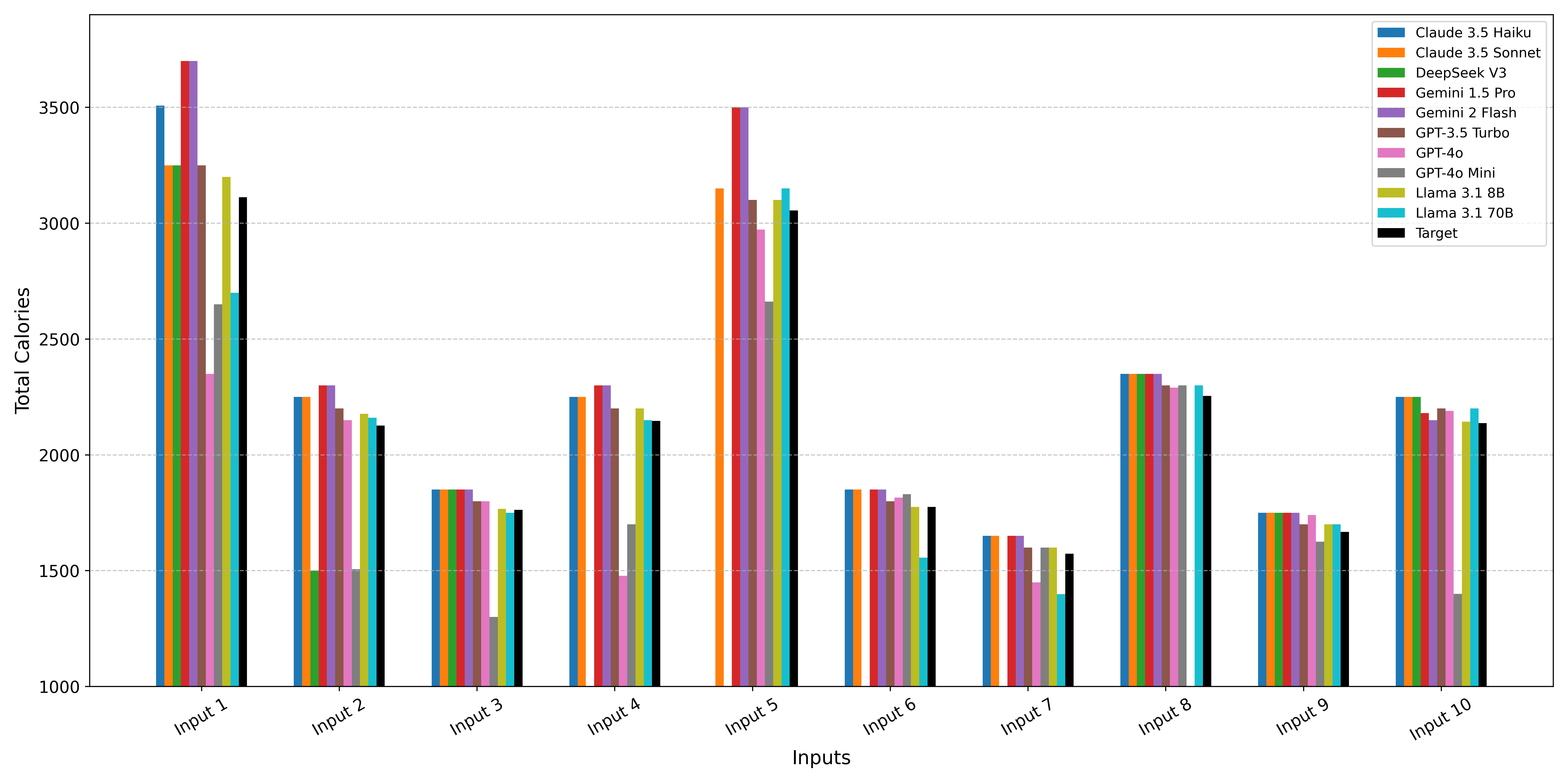}
    \caption{Comparison of the average total calories in meal plans generated by each model against the specified target for each input. The black bars represent the target values, while colored bars indicate the total calories predicted by different models. This visualization highlights the variation in model performance and their ability to generate meal plans that align with user-defined nutritional goals.}
    \label{fig:usda_vs_target}
\end{figure*}

Another valuable analysis that provides insight into how well LLMs incorporate user preferences into their generated meal plans involves comparing the average total nutrition for each input against the specified target values. Fig.~\ref{fig:usda_vs_target} illustrates the comparison between the average total calories in meal plans generated by each model and the target for each input. 

To quantify this difference, we computed the MAE between the actual total calories and the target values, as presented in Table \ref{table:average_error_comparison}. The MAE is calculated as follows:

\begin{equation}
MAE = \frac{1}{N} \sum_{i=1}^{N} \left| \frac{1}{M} \sum_{j=1}^{M} \sum_{k=1}^{P} c_{\text{actual}, i, j, k} - C_{\text{target}, i} \right|
\end{equation}

\noindent where:
\begin{itemize}
    \item \( N = 10 \) represents the total number of inputs.
    \item \( M = 3 \) represents the number of meal plans per input.
    \item \( P \) represents the number of items in each meal plan.
    \item \( c_{\text{actual}, i, j, k} \) is the calorie content of the \( k \)-th food item in the \( j \)-th meal plan for input \( i \).
    \item \( C_{\text{target}, i} \) is the target calorie value for input \( i \).
\end{itemize}
This formulation ensures that the actual total calories for each meal plan are computed as the sum of the calorie content of all individual food items, providing a more detailed and accurate assessment of how closely generated meal plans align with the intended calorie targets.

As shown in Table \ref{table:average_error_comparison}, \textbf{Llama 3.1 8B} and \textbf{GPT-3.5 Turbo} achieved the lowest MAE of 34.14 and 54.16 calories, respectively, with corresponding percentage errors of \textbf{1.55\%} and \textbf{3.68\%}. These results indicate that both models effectively generate meal plans that closely adhere to user-defined caloric targets, with minimal deviation. Meanwhile, GPT-4o Mini exhibited the highest absolute error of 329.06 calories and the largest percentage error of 24.67\%, highlighting substantial discrepancies in its caloric estimations. These findings indicate that while some models effectively minimize deviations in absolute values, others struggle with maintaining proportional accuracy, impacting their reliability in personalized meal planning.

\begin{table}[!tb]
    \centering
    \caption{Mean Absolute and Percentage Error Between Total Calories and Target for Each Model}
    \label{table:average_error_comparison}
    \setlength{\tabcolsep}{1em}
    \renewcommand{\arraystretch}{1.2}
    \begin{tabular}{l|c|c}
        \toprule
        \textbf{Model} & \textbf{MAE} & \textbf{MAE (\%)} \\
        \midrule
        Claude 3.5 Haiku & 128.23 & 8.99 \\
        Claude 3.5 Sonnet & 99.16 & 4.85 \\
        DeepSeek V3 \cite{liu2024deepseek} & 190.61 & 4.85 \\
        Gemini 1.5 Pro \cite{team2023gemini} & 182.16 & 10.44 \\
        Gemini 2 Flash \cite{team2023gemini} & 179.16 & 9.74 \\
        GPT-3.5 Turbo \cite{ye2023comprehensive} & 54.16 & 3.68 \\
        GPT-4o \cite{hurst2024gpt} & 189.76 & 13.47 \\
        GPT-4o Mini & 329.06 & 24.67 \\
        \textbf{Llama 3.1 8B \cite{grattafiori2024llama}} & \textbf{34.14} & \textbf{1.55} \\
        Llama 3.1 70B \cite{grattafiori2024llama} & 109.21 & 8.08 \\
        \bottomrule
    \end{tabular}
\end{table}

\subsection{Evaluation of Meal Plan Quality and Completeness}
Beyond evaluating the nutritional accuracy of generated meal plans, it is equally important to assess whether they adhere to the instructions specified in the prompts. In this analysis, we focus on the structural completeness and richness of the meal plans, including their ability to provide three distinct meal plans, maintain alignment with the personalized nutrition profile, and include detailed recipes.

Our findings indicate that DeepSeek V3 failed to generate meal plans for inputs 4, 5, 6, and 7, while Claude 3.5 Haiku was unable to produce a meal plan for input 5, and Llama 3.1 8B failed to generate a meal plan for input 8. Additionally, Claude 3.5 Haiku produced only one meal plan instead of three for inputs 6, 8, 9, and 10. Another notable issue was observed with Gemini 1.5 Pro, which did not include nutritional values such as fat and protein in any of the generated meal plans, despite these targets being explicitly provided in the input. One possible explanation for why some models, particularly smaller versions, struggled to generate complete meal plans is the constraint on output token length. Each meal plan requires detailed recipes, nutritional values, and food items, resulting in lengthy outputs that may exceed the token limit imposed on smaller models.

It is also worth noting that all models except DeepSeek V3 and GPT-3.5 Turbo included an important disclaimer stating that the provided nutritional values may vary depending on portion sizes and could differ slightly from actual values. This suggests that most models recognize inherent variability in nutritional estimations, contributing to more realistic and informative meal plan generation.

\section{FUTURE WORK}
Despite its strengths, NutriGen has limitations that require further improvement. Some language models encountered output token constraints, leading to incomplete meal plans missing key nutritional details. Inconsistencies in calorie estimation also highlight the need for a more robust retrieval mechanism. Additionally, the use of simulated user data may limit the generalizability of the findings.

To address these issues, future work will explore RAG to dynamically fetch real-time USDA data, reducing retrieval errors and improving nutrient calculations. NutriGen also lacks interactivity, limiting user control over meal plan customization. Integrating a ChatBot interface would allow real-time updates to preferences, restrictions, and nutrition profiles, enhancing engagement and personalization. Further, incorporating multimodal LLMs with food image recognition could improve recommendations by analyzing visual dietary input, assessing ingredient availability, and suggesting alternatives. Although currently limited to English, NutriGen’s language-agnostic design allows adaptation to other languages through localized databases and multilingual LLMs like GPT-3.5 Turbo.

\section{CONCLUSIONS}
This study introduced NutriGen, a personalized meal plan generation framework that leverages LLMs and prompt engineering to bridge the gap between dietary science and practical food recommendation systems. By integrating user-defined dietary preferences, target nutritional goals, and USDA-referenced nutrition data, our approach enables LLMs to generate structured, comprehensive, and nutritionally aligned meal plans. Through rigorous evaluation, we demonstrated that LLMs can effectively generate diverse meal plans while maintaining flexibility, ease of use, and adherence to dietary constraints. Notably, Llama 3.1 8B and GPT-3.5 Turbo achieved the lowest percentage errors of 1.55\% and 3.68\%, respectively, closely matching user-defined caloric targets with minimal deviation. Additionally, GPT-3.5 Turbo exhibited one of the fastest response times, making it a practical choice for real-time meal generation while also demonstrating strong overall performance in minimizing MAE in calorie estimation.

\section{ACKNOWLEDGMENT}
This work was supported in part by the National Science Foundation under grant IIS-2402650. Any opinions, findings, conclusions, or recommendations expressed in this material are those of the authors and do not necessarily reflect the views of the funding organizations.

\bibliographystyle{IEEEtran}
\bibliography{main}

\end{document}